\documentclass[10pt]{article}

\usepackage[margin=1.4in]{geometry}

\usepackage{amsmath}
\usepackage{amssymb}
\usepackage{footnote}
\usepackage[normalem]{ulem}
\makesavenoteenv{tabular}
\makesavenoteenv{table}

\usepackage{graphicx}
\usepackage{color}
\usepackage{subfigure}
\usepackage{amsmath}
\usepackage{amsfonts}
\usepackage{microtype}
\usepackage{graphicx}
\usepackage{subfigure}
\usepackage{booktabs} 
\usepackage{hyperref}
\usepackage{setspace}

\usepackage{pgfplots}
\pgfplotsset{compat=1.14}

\def\X{\mathbf X}

\def\x{\mathbf x}

\def\y{\mathbf y}

\def\ttheta{\boldsymbol \theta}

\def\t{^\top}
\def\eps{\boldsymbol \varepsilon}

\def\N{\mathcal N}

\DeclareMathOperator*{\argmin}{argmin~} 
\DeclareMathOperator{\sign}{sign} 

\newtheorem{theorem}{Theorem}

\newtheorem{property}[theorem]{Property}

\usepackage[round]{natbib}
\renewcommand{\bibname}{References}
\renewcommand{\bibsection}{\subsubsection*{\bibname}}

\usepackage{authblk}

\title{Foothill: A Quasiconvex Regularization for Edge Computing of Deep Neural Networks}
\author[1,2]{Mouloud Belbahri  \thanks{Corresponding author: mouloud.belbahri@huawei.com}}
\author[1]{ Eyy\"ub Sari}
\author[1]{Sajad Darabi}
\author[1]{Vahid Patrovi Nia}
\affil[1]{Huawei Noah's Ark Lab}
\affil[2]{Department of Mathematics and Statistics - University of Montreal}
\date{}

\begin{document}

\maketitle

\begin{abstract}
Deep neural networks (DNNs) have demonstrated success for many supervised learning tasks, ranging from voice recognition,  object detection, to image classification. However, their increasing complexity might yield poor generalization error that make them hard to be deployed on edge devices. Quantization is an effective approach to compress DNNs in order to meet these constraints. Using a quasiconvex base function in order to construct a binary quantizer helps training binary neural networks (BNNs) and adding noise to the input data or using a concrete regularization function helps to improve generalization error. Here we introduce \emph{foothill} function, an infinitely differentiable quasiconvex function. This regularizer is flexible enough to deform towards $L_1$ and $L_2$ penalties. Foothill can be used as a binary quantizer, as a regularizer, or as a loss. In particular, we show this regularizer reduces the accuracy gap between BNNs and their full-precision counterpart for image classification on ImageNet.\\

\noindent \textbf{Copyright Notice:} This paper has been accepted in 16th International Conference of Image Analysis and Recognition (ICIAR 2019). Authors have assigned The Springer Verlage all rights under the Springer copyright. The revised version of the article is published in the Lecture Notes in Computer Science Springer series under the same title.
\end{abstract}

\section{Introduction}
Deep learning has seen a surge in progress, from training shallow networks to very deep networks consisting of tens to hundreds of layers. Deep neural networks (DNNs) have demonstrated success for many supervised learning tasks  \citep{szegedy2015going,simonyan2014very}. The focus has been on increasing accuracy, in particular for image, speech, and recently text tasks, where deep convolutional neural networks (CNNs) are applied. The resulting networks often include millions to billions parameters. Having too many parameters increases the risk of over-fitting and hence a poor model generalization afterall. Furthermore, it is hard to deploy DNNs on low-end edge devices which have tight resource constraints such as memory size, battery life, computation power, etc. The need for models that can operate in resource-constrained environments becomes more and more important.

Quantization is an effective approach to satisfy these constraints. Instead of working with full-precision values to represent the parameters and activations, quantized representations use more compact formats such as integers or binary numbers. Often, binary neural networks (BNNs) are trained with heuristic methods \citep{rastegari2016xnor,hubara2016binarized}. However it is possible to embed the loss function with an appropriate regularization to encourage  binary training \citep{bnnplus}. Common regularizations  encourage the weights to be estimated near zero. Such regularization are not aligned with the objective of training binary networks where the weights are encouraged to be estimated $-1$ or $+1$. Using a regularization function specifically devised for binary quantization \citep{nia2018binary}, it is shown how to modify the objective function in back-propagation to quantize DNNs into one bit with a scaling factor using a quasiconvex base.

In deep learning regularization is sometimes hidden in heuristic methods during training. For instance, adding noise to the input data yields to generalization error improvement \citep{bishop1995training,rifai2011adding}. Data augmentation, and early stopping are some other heuristic regularizations widely applied in practice. A more theoretically sound regularization method is dropout \citep{srivastava2014dropout}, a widely-used method for addressing the problem of over-fitting. The idea is to drop units randomly from the neural network during training. It is known that dropout improves the test accuracy compared to conventional regularizers such as $L_1$ \citep{Tibshirani_lasso_1996} and $L_2$  \citep{srivastava2014dropout}. \cite{wager2013dropout} proved that dropout is equivalent to an $L_2$-type regularizer applied after scaling the inputs.

Linear regression can be regarded as the simplest neural network, with no hidden layer and a linear activation function. Therefore, it is important to study regularization in linear regression context. 


Inspired by the extensive research literature on regularization in the statistical community we introduce \emph{foothill} as a quasiconvex function with attractive properties with strong potentials to be applied in practice in neural network quantization, training neural networks, linear regression, and robust estimation. 

This  function is a generalization of lasso and Ridge penalties and has a strong potential to be used in deep learning. First, we start studying  attractive functional properties of foothill that motivates its use. Then, we demonstrate its application in neural networks binary quantization and neural networks training. Foothill is flexible enough to be used  as a regularizer or even as a loss function. 

\section{Foothill Regularizer}

Let us define the mathematical notation first. Denote univariate variables with lowercase letters, e.g. $x$, vectors with lowercase and bold letters, e.g. $\x$, and matrices with uppercase and bold letters, e.g. $\X$.

\subsection{Definition}

\begin{figure}[htb]
    \centering
    \begin{tikzpicture}[scale=0.5]
    \begin{axis}[
      axis x line=middle, axis y line=middle, samples = 200,
      ylabel={$p_{\alpha,\beta}(x)$},ylabel style={font=\Large, yshift=3pt},
      ymin=-0.1, ymax=2.5, ytick={0,1,2},
      xmin=-3, xmax=3, xtick={-1, 0, 1}
    ]
    \addplot[black, domain=-2.4:2.4, dashed]{1*2*(x-0)*((1+pow(e,-1*(x-0)))^(-1) - 0.5)};
    \addplot[black, domain=-2:2, smooth]{1*2*(x-0)*((1+pow(e,-50*(x-0)))^(-1) - 0.5)};
    \end{axis}
    \end{tikzpicture} 
    \hspace{0.5cm}
    \begin{tikzpicture}[scale=0.5]
            \begin{axis}[
      axis x line=middle, axis y line=middle, samples = 100,  ylabel={$\frac{d^{(l)}p_{1,1}(x)}{d^{(l)}x}$},ylabel style={font=\Large, yshift=3pt}, ymin=-2, ymax=2, ytick={-1,0,1}, xtick={-2.4, -1, 0, 1, 2.4},
      xmin=-4, xmax=4
    ]
    \addplot[black, domain=-4:4, dashed]{2*(1+pow(e,-1*x))^(-1)*(1+1*x*pow(e,-1*x)*(1+pow(e,-1*x))^(-1)))-1};
    \addplot[black, domain=-4:4, smooth]{2 * (-1^2 * x * pow(e, -1*x) * (1+pow(e,-1*x))^(-2) + 2 * 1^2 * x * pow(e, -2*1*x) * (1+pow(e,-1*x))^(-3) + 2 * 1 * pow(e, -1*x) * (1+pow(e,-1*x))^(-2)) };
    \end{axis}
    \end{tikzpicture} 
    \caption{Left panel: Regularization function \eqref{eq:reswish} for $\alpha=1$, $\beta=1$ (solid line)  and $\alpha=1$, $\beta = 50$ (dashed line). Right panel: The first (dashed line) and the second (solid line) derivatives of the regularization function \eqref{eq:reswish} for $\alpha=1$ and $\beta=1$.
    }
    \label{fig:foothill}
\end{figure}
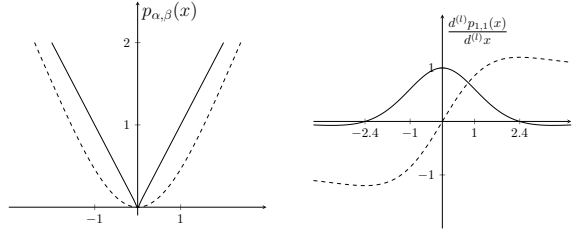

Define the foothill regularization function as

\begin{align}
    p_{\alpha,\beta}(x) = \alpha x~ \mathrm{tanh}\left( \frac{\beta x}{2} \right).
    \label{eq:reswish}
\end{align}

where $\mathrm{tanh}(.)$ is the hyperbolic tangent function, $\alpha > 0$ is a shape parameter and $\beta > 0$ is a scale parameter. 
The function is symmetric about $0$ (see Figure \ref{fig:foothill}, left panel). The first and the second derivatives (see Figure \ref{fig:foothill}, right panel) of foothill are 

\begin{align*}
    \frac{dp_{\alpha,\beta}(x)}{dx} = \alpha ~ \mathrm{tanh}\left( \frac{\beta x}{2} \right) + \frac{1}{2} \alpha \beta x ~ \mathrm{sech}^2\left( \frac{\beta x}{2} \right), \\
     \frac{d^2p_{\alpha,\beta}(x)}{d^2x} = \frac{1}{2} \alpha \beta~ \mathrm{sech}^2\left( \frac{\beta x}{2} \right) \left\{  2 - \beta x ~ \mathrm{tanh}\left( \frac{\beta x}{2} \right) \right\},
\end{align*}
where $\mathrm{sech}(.)$ is the hyperbolic secant function.

\subsection{Properties}

The regularization function (\ref{eq:reswish}) has several interesting properties. It is infinitely differentiable and symmetric about the origin, $$ p_{\alpha,\beta}(x) =  p_{\alpha,\beta}(-x).$$ 
Also, it is flexible enough to approximate the lasso \citep{Tibshirani_lasso_1996} and Ridge penalties \citep{hoerl1970ridge} for particular values of $\alpha$ and $\beta$. The following properties suggest that this function could be considered as a quasiconvex alternative to the elastic net penalty \citep{zou2005regularization}.

\begin{property}  For $\alpha=1$ and $\beta \rightarrow \infty$, the foothill penalty (\ref{eq:reswish}) converges to the lasso penalty.
\end{property}

\emph{Proof.} For $x>0$, it is easy to see that
\begin{flalign*}
    \lim_{\beta \rightarrow +\infty}& \mathrm{tanh}\left( \frac{\beta x}{2} \right) = 1,\\
	\lim_{\beta \rightarrow +\infty}& p_{\alpha, \beta}(x) = x.
\end{flalign*}

Equivalently, for negative $x$, as $p_{\alpha, \beta}(x)$ is symmetric about the origin, then $\lim_{\beta \rightarrow +\infty} p_{\alpha, \beta}(x) = -x$, which is equivalent to $p_{\alpha, \beta}(x) \rightarrow |x|$ when $\beta \rightarrow +\infty$. $\blacksquare$

\begin{property}  For $\alpha>0$, $\beta>0$, and $\beta = 2 / \alpha$ the foothill penalty  (\ref{eq:reswish}) approximates the Ridge penalty in a given interval $[-c ; c]$.
\end{property}

\emph{Proof.} Let us study this property formally. Take the Taylor expansion of (\ref{eq:reswish}), 
\begin{align}
    p_{\alpha, \beta}(x) \approx \frac{\alpha\beta}{2} x^2 - \frac{\alpha\beta^3}{24} x^4 + \frac{\alpha\beta^5}{240} x^6 + \mathrm{O}(x^8).
	\label{eq:taylor}
\end{align}

And, for a given $c>0$, 
\begin{align}
    \int_{0}^{c} \left( \frac{\alpha\beta}{2} x^2 - p_{\alpha, \beta}(x)  \right)^2 dx \approx \frac{\alpha^2\beta^6}{5184} c^9 + \mathrm{O}(c^{11}).
	\label{eq:approxridge}
\end{align}

\begin{figure}[htb]
    \centering
    \includegraphics[scale=0.5]{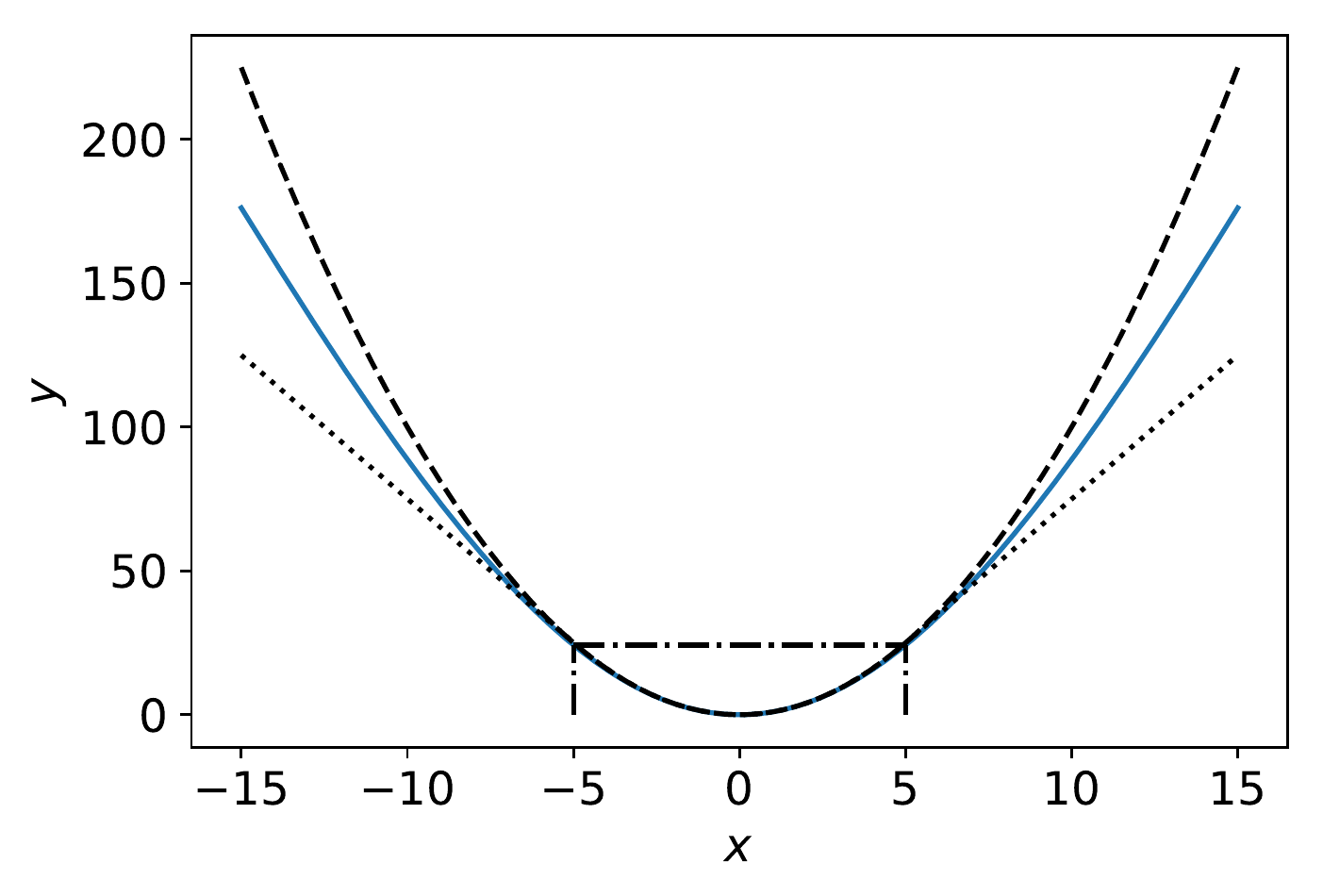} 
    \caption{Plots of the Ridge, the foothill penalty with $\alpha = 16$ and $\beta=0.125$ and twice the Huber loss. The solid blue line represents the foothill penalty, the dashed line represents the Ridge one (upper bound) and Huber (lower bound) is represented in dotted line.}
    \label{fig:approxridge}
\end{figure}

The integral in (\ref{eq:approxridge}) diverges if $c$ tends to infinity, but for a finite positive number $c$, one can numerically estimate the minimal distance between the $L_2$ norm and (\ref{eq:reswish}) with a tiny approximation error. This can be achieved by taking $\beta = 2 / \alpha$ and (\ref{eq:approxridge}) becomes 
\begin{align}
    \int_{0}^{c} \left( x^2 - p_{\alpha, \beta}(x)  \right)^2 dx \approx \frac{1}{81\alpha^4} c^9 + \mathrm{O}(c^{11}) = \varepsilon_c .
	\label{eq:approxridge2}
\end{align}
For large values of $\alpha$, the error $\varepsilon_c$ is negligible, see for example Figure \ref{fig:approxridge} where the regularization function (\ref{eq:reswish}) approximates the Ridge penalty almost perfectly within $[-5 ; 5]$. Furthermore, for fixed parameters, note that 
\begin{align*}
    \lim_{x \rightarrow +\infty} p_{\alpha, \beta}(x) - \alpha x = 0, ~~\mathrm{and}~~  
	\lim_{x \rightarrow -\infty} p_{\alpha, \beta}(x) + \alpha x = 0. ~~ \blacksquare
\end{align*}

Hence it is also interesting to note that (\ref{eq:reswish}) behaves like a polynomial function for small values of $x$, and like a linear function for large values. Therefore, using it as a loss function (instead of a regularization), (\ref{eq:reswish}) behaves like the Huber loss used in practice for robust estimation \citep{huber1964robust}. Figure \ref{fig:approxridge} shows that (\ref{eq:reswish}) is bounded between the Huber loss and the squared error loss.

\begin{property}  Saddle points of $p_{\alpha, \beta}(x)$ are $x_0 \approx \pm 2.3994/\beta$ and $p_{\alpha, \beta}(x_0) = \frac{2 \alpha }{\beta}$.
\end{property}

\emph{Proof.} Indeed, the second order derivative vanishes at
\begin{align*}
     2 - \beta x ~ \mathrm{tanh}\left( \frac{\beta x}{2} \right) = 0,\\
\end{align*}
which is solved by an iterative method for  $\beta x \approx \pm 2.3994$. This implies
\begin{align*}
     \beta x_0 ~ \mathrm{tanh}\left( \frac{\beta x_0}{2} \right) = 2,
\end{align*}
or equivalently
\begin{align*}
	p_{\alpha, \beta}(x_0) = \frac{2 \alpha }{\beta}. ~~\blacksquare
\end{align*}

\begin{property}  The function $p_{\alpha, \beta}(x)$ is quasiconvex.
\end{property}

\emph{Proof.} It is straight forward to show that $p_{\alpha, \beta}(x)$ is decreasing from $-\infty$ to $0$ and increasing from $0$ to $+\infty$ and any monotonic function is quasiconvex (see Figure \ref{fig:foothill}, right panel). $\blacksquare$

\vspace{0.5cm}
Table \ref{tab:relation} suggests that foothill has the flexibility to be used for feature selection regularizer such as the lasso or used only to shrink the estimator in order to prevent over-fitting like the Ridge. Finally, it also can be used as a loss function for robust regression as an alternative to the Huber loss.

\begin{table}[htb]
\begin{center}
\begin{tabular}{l@{\hspace{0.1in}}l@{\hspace{0.1in}}l@{\hspace{0.1in}}l}
  & Shape $\alpha$ & Scale $\beta$ & Function\\
\hline
Lasso        & $1$ & $+\infty$  & $p_{\alpha, \beta}(x)=|x|$\\
Ridge        & $+\infty$ & $2 / \alpha$ & $p_{\alpha, \beta}(x)=x^2$\\
Huber & $< +\infty$ & $2 / \alpha$ &  $p_{\alpha,\beta}(x) = \alpha x~ \mathrm{tanh}\left( \frac{x}{\alpha} \right)$\\
Foothill   & $1$ & $2$ & $p_{\alpha, \beta}(x)=x \mathrm{tanh}(x)$\\
\end{tabular}
\caption{Relationship to other functions} 
\label{tab:relation}
\end{center}
\end{table}

\section{Models}

We start with motivating the use of foothill regularizer for binary quantization. Then, we study some properties in the linear regression context.

\subsection{Binary Quantization}

In BNNs, weights and activations are binarized using the non-differentiable $\sign$ function during the forward pass. It allows to compute dot product using xnor-popcount operations. However, we need to take the derivative of $\sign$ w.r.t. its input, which does not exist. Therefore, a gradient estimator is required \citep{hubara2016binarized}.
The framework of BNN+ \citep{bnnplus} introduces modified $L_1$ and $L_2$ regularizations functions which encourage the weights to concentrate around $\mu \times \{-1; +1\}$, where $\mu$ is a scaling factor. The modified $L_1$ and $L_2$ regularizations are defined as

\begin{align}
    R_{1}(x) &= | |x| -  \mu |, \label{eq:l1bin}\\
    R_{2}(x) &= ( |x| -  \mu )^2. \label{eq:l2bin}
\end{align}

We follow the generalization of \cite{nia2018binary} and modify \eqref{eq:reswish} to construct a shifted regularization function $\Tilde{p}_{\alpha,\beta}(x)$ as
\begin{align}
    \Tilde{p}_{\alpha,\beta}(x) = p_{\alpha,\beta} \big( x -  \mu \sign(x) \big).
    \label{eq:binary}
\end{align}

\begin{figure}[htb]
    \centering
    \begin{tikzpicture}[scale=0.5]
    \begin{axis}[
      axis x line=middle, axis y line=middle, samples = 200,
      ylabel={$\Tilde{p}_{16,0.125}(x)$},ylabel style={yshift=3pt},
      ymin=-0.1, ymax=2.5, ytick={0,1},
      xmin=-3, xmax=3, xtick={-1.5,-0.5, ..., 0.5, 1.5}
    ]
    \addplot[black, domain=-3:0, smooth]{16*2*(x+1.5)*((1+pow(e,-0.125*(x+1.5)))^(-1) - 0.5)};
    \addplot[black, domain=0:3, smooth]{16*2*(x-1.5)*((1+pow(e,-0.125*(x-1.5)))^(-1) - 0.5)};
    \addplot[black, domain=-3:0, dashed]{16*2*(x+0.5)*((1+pow(e,-0.125*(x+0.5)))^(-1) - 0.5)};
    \addplot[black, domain=0:3, dashed]{16*2*(x-0.5)*((1+pow(e,-0.125*(x-0.5)))^(-1) - 0.5)};
    \end{axis}
    \end{tikzpicture}
    \hspace{0.5cm}
    \begin{tikzpicture}[scale=0.5]
    \begin{axis}[
      axis x line=middle, axis y line=middle, samples = 200,
      ylabel={$\Tilde{p}_{0.5,50}(x)$},ylabel style={yshift=3pt},
      ymin=-0.1, ymax=2.5, ytick={0,1},
      xmin=-3, xmax=3, xtick={-1.5,-0.5, ..., 0.5, 1.5}
    ]
    \addplot[black, domain=-3:0, smooth]{0.5*2*(x+1.5)*((1+pow(e,-50*(x+1.5)))^(-1) - 0.5)};
    \addplot[black, domain=0:3, smooth]{0.5*2*(x-1.5)*((1+pow(e,-50*(x-1.5)))^(-1) - 0.5)};
    \addplot[black, domain=-3:0, dashed]{0.5*2*(x+0.5)*((1+pow(e,-50*(x+0.5)))^(-1) - 0.5)};
    \addplot[black, domain=0:3, dashed]{0.5*2*(x-0.5)*((1+pow(e,-50*(x-0.5)))^(-1) - 0.5)};
    \end{axis}
    \end{tikzpicture}
    \caption{Regularization functions for binary networks (\ref{eq:binary}) with $\alpha=16$ and $\beta=0.125$ (left panel) and $\alpha=0.5$ and $\beta=50$ (right panel). Dashed line is $\mu = 0.5$ and solid line is $\mu = 1.5$. The scaling factor $\mu$ is trainable, as a result the regularization function adapts accordingly.}
    \label{fig:binary}
\end{figure}
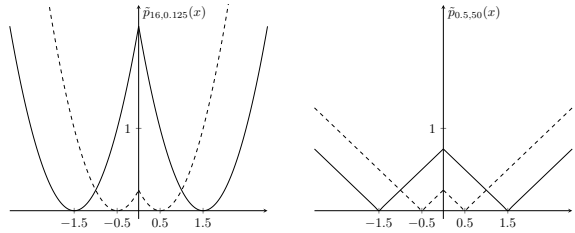

The regularization term is added to the loss function,

$$J(\mathbf{W}, \mathbf{b}) = L(\mathbf{W},\mathbf{b}) + \lambda \sum_{h=1}^H \Tilde{p}_{\alpha, \beta}(\mathbf{W_h}),$$ 

where $L(\mathbf{W}, \mathbf{b})$ is the cost function, $\mathbf{W}$ and $\mathbf{b}$ are the matrices of all weights and bias parameters in the network, $\mathbf{W_h}$ is the matrix of weights at layer $h$ and $H$ is the total number of layers. Here, $\Tilde{p}_{\alpha, \beta}(.)$ is the binary quantizer (\ref{eq:binary}). The regularization function is differentiable, so more convenient to implement in back-propagation. The parameters $\alpha$ and $\beta$ could be defined for the whole network or per layer. In this case, each layer has its own regularization term.

Training the objective function $J(\mathbf{W}, \mathbf{b})$ quantizes the weights around $\pm\mu$ for large values of the regularization constant $\lambda$. Adding the regularization function to the objective function of a deep neural networks adds only one line to the back-propagation in order to estimate the scaling factors. Hence, while training, the regularization function adapts and the weights are encouraged towards $\mu \times \{-1; +1 \}$ (see Figure \ref{fig:binary}). We suggest starting training with $\lambda = 0$ and increasing $\lambda$ with logarithmic rate as a function of the number of epochs \citep{tang2017train,nia2018binary}. The scaling factor and the number of scaling factors are important for BNNs to compete with full-precision networks. In practice, we use a scaling factor per neuron for fully-connected layers and a scaling factor per filter for convolutional layers. Without a scaling factor, the accuracy loss is large \citep{hubara2016binarized}. The scaling factors are applied after the fully-connected and convolutional layers which are performed using xnor-popcount operations during inference. In our experiments, we learn the scaling factors with back-propagation.


\subsection{Regression}
Suppose the response variable is measured with an additive statistical error $\varepsilon$ and the relationship between the response and the predictors is fully determined by a linear function

\begin{align}
	\y=\X \ttheta+\eps,
	\label{eq:lm}
\end{align}
where $\y_{n \times 1}$ is the vector of observed response, $\X_{n \times p}$ is row-wise stacked matrix of predictors, $\ttheta_{p \times 1}$ is the $p$-dimensional vector of coefficients, and $\eps_{n \times 1}$ is white noise with zero mean and a constant variance $\tau^2$. 

The penalized estimator with squared-loss function is defined as

\begin{align}
	\hat{\ttheta} = \argmin_{\ttheta} \frac{1}{2n} \left\lVert \y - \X\ttheta \right\rVert_2^{2} + \lambda \sum_{j=1}^p p_{\alpha, \beta}(\theta_j),
	\label{eq:estimator}
\end{align}

where $p_{\alpha, \beta}(.)$ is the regularization function (\ref{eq:reswish}). Here, $\lambda$ is the regularization constant. Setting $\lambda = 0$ returns the ordinary least squares estimates, which performs no shrinking and no selection. For a given $\lambda > 0$ and finite $\alpha$ and $\beta$, the regression coefficients $\hat \ttheta$ are shrunk towards zero, and for $\alpha =1$, when $\beta \rightarrow +\infty$, (\ref{eq:reswish}) converges to the lasso penalty which sets some of the coefficients to zero (sparse selection), so does selection and shrinkage simultaneously.

To better understand the proposed penalty, we consider the orthogonal case where we assume that the columns of $\X$ in (\ref{eq:lm}) are orthonormal, i.e. $\X^{\t}\X=n\mathbf I_p$. Therefore, the minimization problem of (\ref{eq:estimator}) is equivalent to estimating coefficients component-wise. Let $\hat z_j = \x_j^{\t} \y / n$ be the ordinary least squares estimate for $j=1,...,p$. Here, for fixed $\alpha > 0$ and a given scale parameter $\beta > 0$, this leads us to the univariate optimization problem

\begin{align}
	\argmin_{\theta_j} \left[ \frac{1}{2} (\hat z_j - \theta_j)^2 + \lambda  \alpha \theta_j ~ \mathrm{tanh}\left( \frac{\beta\theta_j}{2}\right) \right].
	\label{eq:orthogonal}
\end{align}

\begin{figure}[ht]
    \centering
    \includegraphics[scale=0.5]{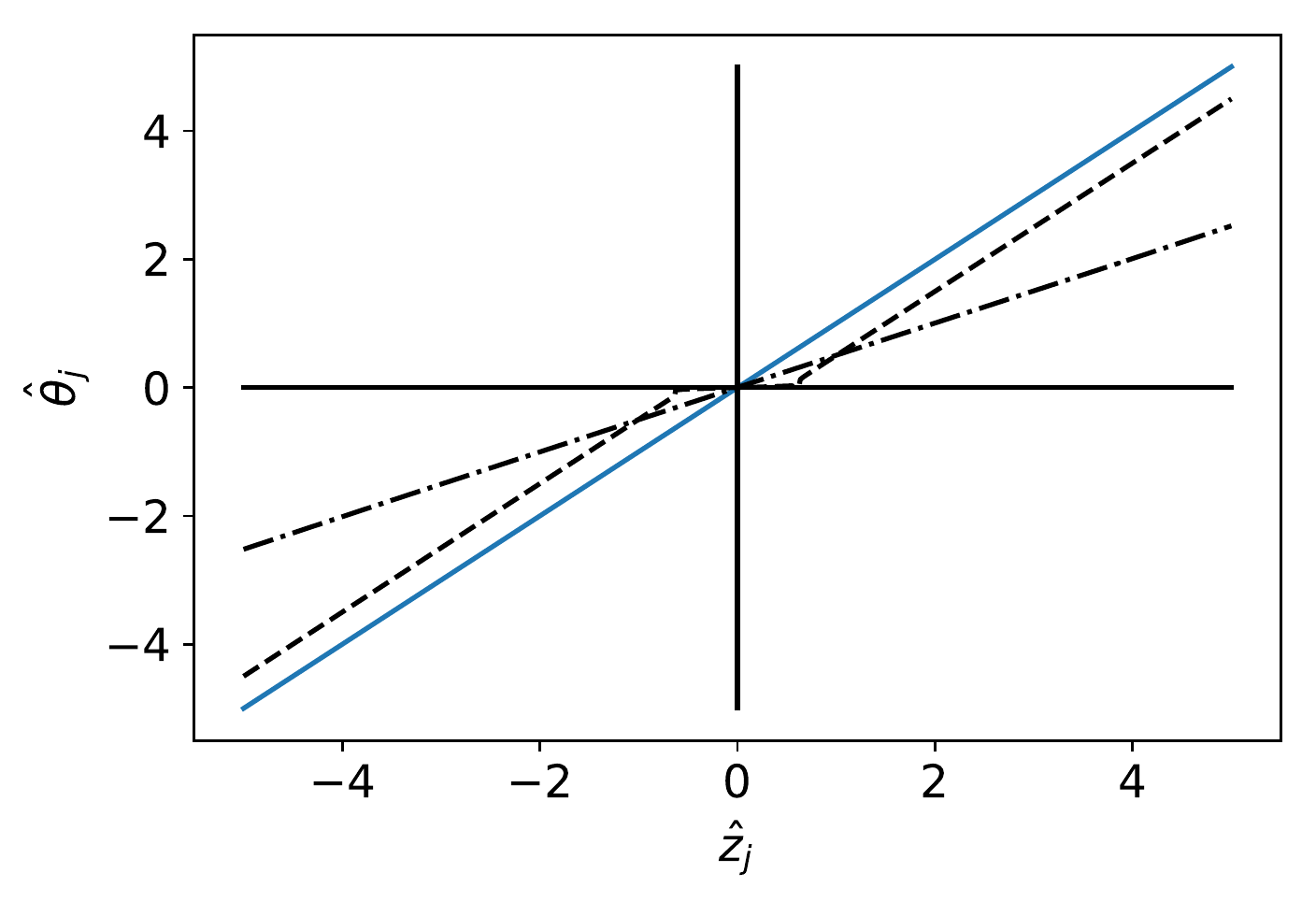} 
    \caption{Solution paths in the orthogonal design study according to the OLS estimator $\hat z_j$ for the foothill with $\alpha=16$ and $\beta=0.125$ (dashed and dotted line) and $\alpha=1$ and $\beta = 50$ (dashed line), with $\lambda =0.5$ . The solid blue line represents the OLS estimator.}
    \label{fig:path}
\end{figure}

The numerical solutions of (\ref{eq:orthogonal}) with various values of $\alpha$ and $\beta$ are shown in Figure \ref{fig:path}. When $\beta$ is small, the solutions are smooth and by increasing $\beta$, the solutions become similar to ones of lasso.

Following \cite{knight2000asymptotics} proof for Bridge regression \citep{frank1993statistical}, we show  that under similar conditions and a fixed $\lambda$, the penalized estimator is $\sqrt{n}$-consistent.

Consider the linear model (\ref{eq:lm}) and denote the penalized least squares function by 

\begin{align*}
    J_{n}(\ttheta) = \frac{1}{2}(\y - \X\ttheta)^{\top}(\y - \X\ttheta) + \lambda \sum_{j=1}^p p_{\alpha, \beta}(\theta_j).
\end{align*}

\begin{property}  Assume that the matrix $\mathbb{E}[\X^{\top}\X] < \infty$ is positive definite. Let $\hat{\ttheta}_n$ be the penalized estimator. $\hat{\ttheta}_n$ is consistent if any given $\epsilon > 0$, there exist a large constant $C$ such that
\begin{align}
    \mathrm{Pr} \left( \inf_{ \left\lVert \mathbf u \right\rVert = C} J_{n}\left(\ttheta + \frac{\mathbf u}{\sqrt{n}}\right) > J_{n}(\ttheta) \right) \geq 1 - \epsilon.
    \label{eq:inf}
\end{align}
\end{property}

\emph{Proof.} This implies there is a local minimizer such that $\lVert \hat{\ttheta}_n - \ttheta \rVert = O_P(\sqrt{n})$. Simple algebra shows that 
\begin{align*}
    D_{n}(\mathbf u) &:= J_{n}(\ttheta + \frac{\mathbf u}{\sqrt{n}}) - J_{n}(\ttheta)\\
    &= \frac{1}{2} \mathbf{u}^{\top} \frac{\X^{\top}\X}{n} \mathbf{u} - \mathbf{u}^{\top} \frac{\X^{\top}(\y - \X\ttheta)}{\sqrt{n}} \\
    &+ \lambda \sum_{j=1}^p \left( p_{\alpha, \beta}\left(\theta_j + \frac{u_j}{\sqrt{n}}\right) - p_{\alpha, \beta} (\theta_j) \right),
\end{align*}

which is minimized at $\sqrt{n} (\hat{\ttheta}_n - \ttheta)$. By the strong law of large numbers and the central limit theorem, the first two terms converge to 
\begin{align*}
    \frac{1}{2} \mathbf{u}^{\top}\mathbb{E}[\X^{\top}\X]\mathbf{u} - \mathbf{u}^{\top} \mathbf{Z},
\end{align*}
where $\mathbf{Z} \sim \N(\mathbf{0}, \boldsymbol{\Sigma})$ where $\boldsymbol{\Sigma} = \tau^2 \mathbb{E}[\X^{\top}\X]$. The third term can be rewritten as 
\begin{align*}
    \frac{\lambda}{\sqrt{n}} \sum_{j=1}^p \left( \frac{p_{\alpha, \beta}\left(\theta_j + \frac{u_j}{\sqrt{n}}\right) - p_{\alpha, \beta} (\theta_j)}{\frac{u_j}{\sqrt{n}}} \right) u_j   , 
\end{align*}
and suppose that $\frac{\lambda}{\sqrt{n}} \rightarrow \lambda_0$. Therefore, when $n \rightarrow +\infty$, we have $\frac{u_j}{\sqrt{n}} \rightarrow 0$ so the third term of $D_{n}$ converges to
\begin{align*}
    \lambda_0 \sum_{j=1}^p \left( \frac{d p_{\alpha, \beta} (\theta_j)}{d \theta_j}  \right) u_j. 
\end{align*}
For $\lambda$ fixed, $\lambda_0 = 0$ and the first derivative of the regularization function is bounded, which means that $D_{n}(\mathbf u)$ converges to 
\begin{align*}
    D(\mathbf u) = \frac{1}{2\tau^2} \mathbf{u}^{\top}\boldsymbol{\Sigma}\mathbf{u} - \mathbf{u}^{\top} \mathbf{Z},
\end{align*}
which is convex and has a unique minimizer and hence,
\begin{align*}
    \sqrt{n} (\hat{\ttheta}_n - \ttheta) \rightarrow_d \argmin D(\mathbf u),
\end{align*}
which shows that by choosing sufficiently large $C$, (\ref{eq:inf}) holds and that $\hat{\ttheta}_n$ is $\sqrt{n}$-consistent. $\blacksquare$

\section{Application}

In this section, we evaluate the performance of foothill on different applications. With two extra parameters, foothill is more flexible than $L_1$ and $L_2$. We believe that its flexibility helps fine-tuning.

\subsection{Binary Quantization}

In this section, we evaluate foothill's performance on a hard task. We use the shifted version from equation (\ref{eq:binary}) in order to quantize a neural network. We quantize AlexNet architecture \citep{krizhevsky2012imagenet} on ImageNet. This dataset consists of $\sim$1.2M training images, $50$K validation images and $1000$ classes. During training, images are resized to $256\times256$ and a random crop is applied to obtain $224\times224$ input size. Random horizontal flip is also used as a data augmentation technique. At test time, images are resized to $256\times256$ and a center crop is applied to get $224\times224$ size. For both steps, standardization is applied with $\mathrm{mean}=[0.485, 0.456, 0.406]$ and $\mathrm{std}=[0.229, 0.224, 0.225]$. Note that AlexNet architecture used for training BNNs is slightly modified from the original architecture as we need to change the order of some operation. For instance, pooling should not be performed after the binary activations. Therefore, we adopt the architecture described in \cite{bnnplus} where batch normalization layers are added \citep{ioffe2015batchnorm}. Weights and activations are quantized using the $\sign$ function for all convolutional and fully-connected layers except the first and the last ones which are kept to be in full-precision. We initialize the learning rate with $5 \times 10^{-3}$ and divide it each $10$ epochs alternatively, by $5$ and by $2$. We use $\lambda = 10^{-6} \times \mathrm{log}(t)$ where $t$ is the current epoch and train the networks for $100$ epochs. We compare our method to traditional binary networks.

\begin{table}[htb]
\begin{center}
\begin{tabular}{lcc}
\multicolumn{1}{c}{Method}  &\multicolumn{1}{c}{Top-1 Accuracy} &\multicolumn{1}{c}{Top-5 Accuracy}
\\ \hline
$R_1(x)$ & $43.0\%$ & $67.5\%$ \\
$\Tilde{p}_{0.5, 50}(x)$ & $44.4\%$ & $68.5\%$ \\ 
$\Tilde{p}_{0.75, 50}(x)$ & $44.3\%$ & $68.4\%$ \\ 
$R_2(x)$ & $42.9\%$ & $67.5\%$ \\
$\Tilde{p}_{100, 0.02}(x)$ & $44.2\%$ & $68.5\%$ \\ 
$\Tilde{p}_{20, 0.1}(x)$ & $\bf 44.5\%$ & $68.3\%$ \\ 
\\
BinaryNet & $41.2\%$ & $65.6\%$  \\
XNOR-Net & $44.2\%$ & $69.2\%$  \\
Full-Precision & $57.1\%$ & $80.3\%$  
\end{tabular}
\caption{Comparison of top-1 and top-5 accuracies of quantized neural network using the lasso (\ref{eq:l1bin}), Ridge (\ref{eq:l2bin}) and foothill (\ref{eq:binary}) modified regularizers to traditional BinaryNet \citep{hubara2016binarized} and XNOR-Net \citep{rastegari2016xnor} on ImageNet dataset, using AlexNet architecture.}
\label{tab:imagenet}
\end{center}
\end{table}

In Table \ref{tab:imagenet}, we report XNOR-Net performance from the original paper of \cite{rastegari2016xnor} and the BinaryNet one from the implementation of \cite{lin2017towards}, which is higher than the one reported in the original paper. We do not report the performance of \cite{bnnplus} as they make use of a pre-trained model in their experiments, whereas we train the binary neural networks from scratch. We see that quantizing a neural network using foothill function as a regularization that pushes the weights towards binary values gives more accurate results for ImageNet dataset, better than $L_1$ and $L_2$ by more than $1.5\%$, which is a big gain for BNNs. Furthermore, for AlexNet architecture, our method beats the state of the art BinaryNet and XNOR-Net.

\subsection{Regularization}

We use AlexNet architecture augmented by batch normalization in order to compare foothill (\ref{eq:reswish}) to $L_1$ and $L_2$ regularizers on CIFAR-10. We train the network for $50$ epochs using stochastic gradient descent optimizer with momentum $0.9$ and a learning rate of $10^{-2}$ that is divided by $10$ at epochs 20 and 30. The data preprocessing pipeline is the same as for ImageNet. For each experiment, the regularization constant $\lambda$ is set to a value in $\{10^{-4}, 10^{-3}, 10^{-2}\}$.

\begin{table}[htb]
\begin{center}
\begin{tabular}{l@{\hspace{0.2in}}ccc@{\hspace{0.2in}}ccc}
\multicolumn{1}{c}{$\lambda$}  
&\multicolumn{1}{c}{$L_1(x)$}  
&\multicolumn{1}{c}{$p_{0.5, 50}(x)$}   
&\multicolumn{1}{c}{$p_{0.75, 50}(x)$}
&\multicolumn{1}{c}{$L_2(x)$}
&\multicolumn{1}{c}{$p_{16, 0.125}(x)$}
&\multicolumn{1}{c}{$p_{20, 0.1}(x)$}
\\ \hline

$10^{-4}$ & $89.75\%$ & $89.53\%$ & $\bf 90.24\%$ & $88.61\%$ & $88.73\%$ & $89.12\%$ \\
$10^{-3}$ & $81.74\%$ & $\bf 90.55\%$ & $90.05\%$ & $89.51\%$ & $89.30\%$ & $89.44\%$ \\
$10^{-2}$ & $55.59\%$ & $85.60\%$ & $84.78\%$ & $89.99\%$ & $89.86\%$ & $\bf 90.21\%$

\end{tabular}
\caption{Regularized AlexNet top-1 accuracies on CIFAR-10 test set, using different $\lambda$ values. Our implementation of the non-regularized AlexNet achieves $88.63\%$ accuracy.} 
\label{tab:cifarcomparison}
\end{center}
\end{table}

\begin{figure}[ht]
    \centering
    \includegraphics[scale=0.5]{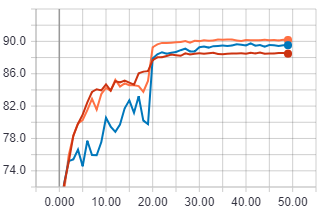} 
    \includegraphics[scale=0.5]{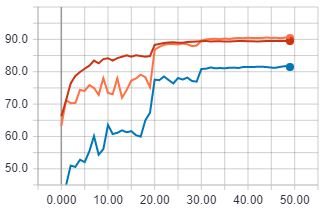}
    \includegraphics[scale=0.5]{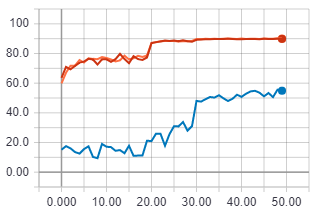}
    \caption{From left to right, $\lambda = 10^{-4}, 10^{-3}, 10^{-2}$ validation curves for $L_1$ (blue), $L_2$ (red) and the best foothill regularizer (orange). The validation curves show the robustness of foothill in comparison with $L_1$ with respect to $\lambda$.}
    \label{fig:val_curves}
\end{figure}

The results reported in Table \ref{tab:cifarcomparison} and Figure \ref{fig:val_curves} empirically demonstrate the flexibility of foothill against $L_1$ and $L_2$. Our regularization function is less sensitive to the choice of $\lambda$. For instance, $L_1$-regularized AlexNet's accuracy can have $34.16\%$ difference depending on which $\lambda$ has been used for training while foothill with $\alpha=0.5$ and $\beta=50$ regularized AlexNet's accuracy difference ranges in $4.96\%$.

\section{Conclusion}
Here we developed a new function, called foothill, that can be used as a binary quantizer, as a regularizer, or a loss function.  

Most of the deep networks includes millions of parameters that requires extensive resources to be implemented in realtime. A modified version of foothill can be used to quantize deep networks and ultimately run neural networks to low power edge devices, such as wearable devices, cell phones, wireless base stations, etc.  Network quantization yields to accuracy degradation. Recent studies \citep{hubara2016binarized,rastegari2016xnor} suggest proper training of weights controls the  accuracy loss. The shift version of foothill has the  potential of pushing heuristic training towards a more clear and formalized training using regularization. Our numerical results confirm this assumption since our implementation of a quantized neural network using foothill regularizer beats $L_1$ and $L_2$ regularizers and XNOR-Net, which is the state of the art binary quantization method. 

As a regularizer foothill  may encourage estimation shrinkage, sparse selection, or both depending on the values of its parameters. More concretely its parameters can be tuned to  approximate both  lasso (which implements sparse selection) and Ridge penalty (which implements shrinkage). Therefore foothill looks like a quasiconvex version of the elastic net which approximates the lasso and the Ridge. As a loss function, the behaviour of foothill is similar to the Huber loss. 

\subsubsection*{Acknowledgements}
We want to acknowledge technical discussions with Matthieu  Courbariaux, Mohan Liu, and Alejandro Murua. This research was not possible without continuous support of Yanhui Geng and Li Zhou. 

\bibliography{main}

\begin{thebibliography}{}

\bibitem[\protect\astroncite{Bishop}{1995}]{bishop1995training}
Bishop, C.~M.\leavevmode\nopagebreak\newline 1995.
\newblock Training with noise is equivalent to tikhonov regularization.
\newblock {\em Neural Computation}, 7(1):108--116.

\bibitem[\protect\astroncite{Darabi et~al.}{2018}]{bnnplus}
Darabi, S., M.~Belbahri, M.~Courbariaux, and V.~P.
  Nia\leavevmode\nopagebreak\newline 2018.
\newblock {BNN+}: Improved binary network training.
\newblock {\em arXiv preprint arXiv:1812.11800}.

\bibitem[\protect\astroncite{Frank and Friedman}{1993}]{frank1993statistical}
Frank, L.~E. and J.~H. Friedman\leavevmode\nopagebreak\newline 1993.
\newblock A statistical view of some chemometrics regression tools.
\newblock {\em Technometrics}, 35(2):109--135.

\bibitem[\protect\astroncite{Hoerl and Kennard}{1970}]{hoerl1970ridge}
Hoerl, A.~E. and R.~W. Kennard\leavevmode\nopagebreak\newline 1970.
\newblock Ridge regression: Biased estimation for nonorthogonal problems.
\newblock {\em Technometrics}, 12(1):55--67.

\bibitem[\protect\astroncite{Hubara et~al.}{2016}]{hubara2016binarized}
Hubara, I., M.~Courbariaux, D.~Soudry, R.~El-Yaniv, and
  Y.~Bengio\leavevmode\nopagebreak\newline 2016.
\newblock Binarized neural networks.
\newblock In {\em Advances in Neural Information Processing Systems}.

\bibitem[\protect\astroncite{Huber et~al.}{1964}]{huber1964robust}
Huber, P.~J. et~al.\leavevmode\nopagebreak\newline 1964.
\newblock Robust estimation of a location parameter.
\newblock {\em The Annals of Mathematical Statistics}, 35(1):73--101.

\bibitem[\protect\astroncite{Ioffe and Szegedy}{2015}]{ioffe2015batchnorm}
Ioffe, S. and C.~Szegedy\leavevmode\nopagebreak\newline 2015.
\newblock Batch normalization: Accelerating deep network training by reducing
  internal covariate shift.
\newblock {\em arXiv preprint arXiv:1502.03167}.

\bibitem[\protect\astroncite{Knight and Fu}{2000}]{knight2000asymptotics}
Knight, K. and W.~Fu\leavevmode\nopagebreak\newline 2000.
\newblock Asymptotics for lasso-type estimators.
\newblock {\em Annals of Statistics}, Pp.~ 1356--1378.

\bibitem[\protect\astroncite{Krizhevsky et~al.}{2012}]{krizhevsky2012imagenet}
Krizhevsky, A., I.~Sutskever, and G.~E. Hinton\leavevmode\nopagebreak\newline
  2012.
\newblock Imagenet classification with deep convolutional neural networks.
\newblock In {\em Advances in Neural Information Processing Systems}, Pp.~
  1097--1105.

\bibitem[\protect\astroncite{Lin et~al.}{2017}]{lin2017towards}
Lin, X., C.~Zhao, and W.~Pan\leavevmode\nopagebreak\newline 2017.
\newblock Towards accurate binary convolutional neural network.
\newblock In {\em Advances in Neural Information Processing Systems}, Pp.~
  345--353.

\bibitem[\protect\astroncite{Nia and Belbahri}{2018}]{nia2018binary}
Nia, V.~P. and M.~Belbahri\leavevmode\nopagebreak\newline 2018.
\newblock Binary quantizer.
\newblock {\em Journal of Computational Vision and Imaging Systems}, 4(1):3--3.

\bibitem[\protect\astroncite{Rastegari et~al.}{2016}]{rastegari2016xnor}
Rastegari, M., V.~Ordonez, J.~Redmon, and
  A.~Farhadi\leavevmode\nopagebreak\newline 2016.
\newblock Xnor-net: Imagenet classification using binary convolutional neural
  networks.
\newblock In {\em European Conference on Computer Vision}, Pp.~ 525--542.
  Springer.

\bibitem[\protect\astroncite{Rifai et~al.}{2011}]{rifai2011adding}
Rifai, S., X.~Glorot, Y.~Bengio, and P.~Vincent\leavevmode\nopagebreak\newline
  2011.
\newblock Adding noise to the input of a model trained with a regularized
  objective.
\newblock {\em arXiv preprint arXiv:1104.3250}.

\bibitem[\protect\astroncite{Simonyan and Zisserman}{2014}]{simonyan2014very}
Simonyan, K. and A.~Zisserman\leavevmode\nopagebreak\newline 2014.
\newblock Very deep convolutional networks for large-scale image recognition.
\newblock {\em arXiv preprint arXiv:1409.1556}.

\bibitem[\protect\astroncite{Srivastava et~al.}{2014}]{srivastava2014dropout}
Srivastava, N., G.~Hinton, A.~Krizhevsky, I.~Sutskever, and
  R.~Salakhutdinov\leavevmode\nopagebreak\newline 2014.
\newblock Dropout: a simple way to prevent neural networks from overfitting.
\newblock {\em The Journal of Machine Learning Research}, 15(1):1929--1958.

\bibitem[\protect\astroncite{Szegedy et~al.}{2015}]{szegedy2015going}
Szegedy, C., W.~Liu, Y.~Jia, P.~Sermanet, S.~Reed, D.~Anguelov, D.~Erhan,
  V.~Vanhoucke, and A.~Rabinovich\leavevmode\nopagebreak\newline 2015.
\newblock Going deeper with convolutions.
\newblock In {\em Proceedings of the IEEE conference on computer vision and
  pattern recognition}, Pp.~ 1--9.

\bibitem[\protect\astroncite{Tang et~al.}{2017}]{tang2017train}
Tang, W., G.~Hua, and L.~Wang\leavevmode\nopagebreak\newline 2017.
\newblock How to train a compact binary neural network with high accuracy?
\newblock In {\em Thirty-First AAAI Conference on Artificial Intelligence}.

\bibitem[\protect\astroncite{Tibshirani}{1996}]{Tibshirani_lasso_1996}
Tibshirani, R.\leavevmode\nopagebreak\newline 1996.
\newblock Regression shrinkage and selection via the lasso.
\newblock {\em Journal of the Royal Statistical Society: Series B}, Pp.~
  267--288.

\bibitem[\protect\astroncite{Wager et~al.}{2013}]{wager2013dropout}
Wager, S., S.~Wang, and P.~S. Liang\leavevmode\nopagebreak\newline 2013.
\newblock Dropout training as adaptive regularization.
\newblock In {\em Advances in Neural Information Processing Systems}, Pp.~
  351--359.

\bibitem[\protect\astroncite{Zou and Hastie}{2005}]{zou2005regularization}
Zou, H. and T.~Hastie\leavevmode\nopagebreak\newline 2005.
\newblock Regularization and variable selection via the elastic net.
\newblock {\em Journal of the Royal Statistical Society: Series B},
  67(2):301--320.

\end{thebibliography}
\end{document}